\documentclass[journal]{IEEEtran}
\usepackage{amsmath,amsfonts}
\usepackage{algorithmic}
\usepackage{array}
\usepackage[caption=false,font=normalsize,labelfont=sf,textfont=sf]{subfig}
\usepackage{textcomp}
\usepackage{stfloats}
\usepackage{url}
\usepackage{verbatim}
\usepackage{graphicx}
\usepackage{amsmath,amssymb} 
\usepackage{multirow}
\usepackage{color}
\def\BibTeX{{\rm B\kern-.05em{\sc i\kern-.025em b}\kern-.08em
    T\kern-.1667em\lower.7ex\hbox{E}\kern-.125emX}}
\usepackage{balance}
\begin{document}
\title{Generative Model-Based Feature Attention Module for Video Action Analysis}
% \author{Guiqin Wang, Peng Zhao, Cong Zhao, Jing Huang, Siyan Guo, Shusen Yang}

\author{Guiqin Wang , Peng Zhao
\IEEEmembership{Member, IEEE}, Cong Zhao, Jing Huang, Siyan Guo, Shusen Yang
\IEEEmembership{Senior Member, IEEE}
\thanks{G. Wang, P. Zhao (Corresponding Author) and J. Huang are with the School of Computer Science and
Technology, and with the National Engineering Laboratory for Big Data Analytics (NEL-BDA), Xi’an Jiaotong University, Xi’an 710049, China (e-mail: gqwang@stu.xjtu.edu.cn; p.zhao@mail.xjtu.edu.cn; Huang.J@stu.xjtu.edu.cn).}
\thanks{C. Zhao  and S. Guo are with the School of Mathematics and Statistics, and with the the National Engineering Laboratory for Big Data
Analytics (NEL-BDA), Xi’an Jiaotong University, Xi’an 710049, China (e-mail: congzhao@xjtu.edu.cn; guosiyan813@stu.xjtu.edu.cn).}
\thanks{S. Yang is with the School of Mathematics and Statistics, with the National Engineering Laboratory for Big Data Analytics (NEL-BDA), and with the Ministry of Education Key Lab for Intelligent Networks and Network Security (MOE KLINNS Lab), Xi’an Jiaotong University,
Xi’an, Shaanxi 710049, China; and also with Pazhou Laboratory, Guangzhou,
Guangdong 510335, China (email: shusenyang@mail.xjtu.edu.cn).}
}

\markboth{Journal of \LaTeX\ Class Files,~Vol.~18, No.~9, September~2020}%
{How to Use the IEEEtran \LaTeX \ Templates}

\maketitle

\begin{abstract}
  Video action analysis is a foundational technology within the realm of intelligent video comprehension, particularly concerning its application in Internet of Things(IoT). However, existing methodologies overlook feature semantics in feature extraction and focus on optimizing action proposals, thus these solutions are unsuitable for widespread adoption in high-performance IoT applications due to the limitations in precision, such as autonomous driving, which necessitate robust and scalable intelligent video analytics analysis. To address this issue, we propose a novel generative attention-based model to learn the relation of feature semantics. Specifically, by leveraging the differences of actions' foreground and background, our model simultaneously learns the frame- and segment-dependencies of temporal action feature semantics, which takes advantage of feature semantics in the feature extraction effectively. To evaluate the effectiveness of our model, we conduct extensive experiments on two benchmark video task, action recognition and action detection. In the context of action detection tasks, we substantiate the superiority of our approach through comprehensive validation on widely recognized datasets. Moreover, we extend the validation of the effectiveness of our proposed method to a broader task, video action recognition. Our code is available at \url{https://github.com/Generative-Feature-Model/GAF}.
\end{abstract}


\section{Introduction}

\IEEEPARstart{I}{n} tandem with the unprecedented proliferation of Internet of Things (IoT) visual sensors today, encompassing urban surveillance cameras, cameras integrated within autonomous vehicles, and those embedded in smartphones, there has been a remarkable surge in the volume of video data~\cite{chen2020internet}. To effectively manage and analyze this influx of video data, numerous researchers in both academia and industry have turned their attention to intelligent video analysis, a versatile technique applicable across diverse domains such as traffic management~\cite{ni2024integrated}, highlight extraction~\cite{zhang2023cross}, and sports analysis~\cite{wu2022sports}. Of particular significance, as a fundamental task within intelligent video analysis, \textit{video action analysis} is dedicated to the recognition and detection of human actions within videos of varying lengths and uncertainty.

% \IEEEPARstart{A}{ccompanying} the unprecedented adoption of Internet of Things(IoT) visual sensors today, including urban surveillance cameras, cameras within autonomous vehicles, and those incorporated into smartphones, is a remarkable escalation in the quantity of video data~\cite{chen2020internet}. To efficiently analyze this video data, numerous researchers in both academia and industry focus on intelligent video analysis, a versatile technique applicable across various domains such as traffic management~\cite{ouallane2022fusion}, highlight extraction~\cite{zhang2023cross}, and sports analysis~\cite{wu2022sports}.
% Specially, as a foundational task within intelligent video analysis, \textit{video action analysis} is dedicated to the recognition and detection of human action within videos of varying lengths and uncertainty.
% \IEEEPARstart{I}{n} recent years, with the tremendous increase in electronic equipment, a large number of videos are recorded and uploaded from daily activities. Temporal Action Detection(TAD), aiming to detect the action boundaries(\textit{i.e.}, the start and the end time) and the category label of action instance, is a fundamental task of intelligent video analysis. It can localize useful video fragments from a long untrimmed video, facilitating applications(\textit{e.g.}, smart surveillance, sports analysis, highlight extraction)~\cite{kong2018human,ren2018learning,liu2018com}.

In exploring the domain of video action analysis, researchers predominantly focus on two pivotal tasks: video action recognition~\cite{Wang2021tdn,ouallane2022fusion,liu2022adver,hussain2020multiview} and video action detection~\cite{gao2019video,lin2019bmn,liu2019multi}. For enhancing the ability of analyzing video action, existing methods devise various backbone for capturing video action features, such as 3D-CNN~\cite{ji20123d}, two-stream~\cite{simonyan2014two} and Transformer-based~\cite{yang2022recurring} methods. Simultaneously, certain researchers are investigating attention modules as a means to enhance the performance of established backbone architectures(\textit{e.g.}, graph-attention~\cite{liu2022graph}, relation-attention~\cite{chen2019relation}, class-semantic-attention~\cite{sridhar2021class}). Remarkably, while Transformer-based attention mechanisms represent an efficient approach, their resource-intensive nature poses limitations for deployment on resource-constrained devices, particularly in the context of video tasks~\cite{wang2023generative}.

% TAD is related to object detection in image analysis because both of them try to propose meaningful regions(2-D spatial region and 1-D temporal intervals)
% and then detect them. Inspired by the success of object detection, most TAD methods can be divided into two categories, action localization and action classification. 
% With contemporary action classifiers~\cite{Wang2021tdn,Chen2021deep,liu2022adver} achieving compelling performance, the action 
% classification achieves cogent performance. To improve the performance of TAD on standard benchmarks, precisely locating the 
% action instances remains a challenge. Many early studies use hand-craft features for action localization, such as 
% DLSBP~\cite{duchenne2009automatic}, ASM~\cite{gaidon2011actom} and SDPM~\cite{tian2013spatiotemporal}. In the past decade, due to the 
% success of deep learning, most current methods mainly focus on CNN-based feature extraction of action localization~\cite{gao2019video,lin2019bmn,liu2019multi}.

\begin{figure}
  \centering
  \includegraphics[width=0.45\textwidth]{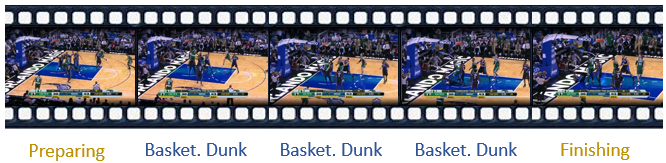}
  \caption{Feature semantics of small objectives and complex background are necessary to locate action instances in this video.}
  \label{fig:0}
\end{figure}

Nonetheless, existing video action analysis~\cite{xing2023svformer,gao2019video,lin2019bmn} methods mainly focus on the optimization of the action region and modelling of the action relation. In other word, current methods is aiming at learning the segment-level information of the video action. Despite the successful progress of this solution, the majority of methodologies heavily depend on the modelling of action segments, with the objective of optimizing feature extraction as the input for action-related tasks, such as temporal action classification and temporal action detection. In practical applications, both segment-level features and frame-level features of a video play crucial roles: segment-level features are valuable for modeling the relations between action segments~\cite{wu2019long,chen2019relation,zhnag2020ada}, while frame-level features are instrumental in accurately locating action objectives, particularly in the presence of complex background elements~\cite{quader2020weight,wang2024paxion}.
% The intra-relations inside each feature are effective for adjusting for the wrong action labels, and the inter-relations across different features are helpful for correcting the imprecise action boundaries~\cite{zhnag2020ada,zheng2021coll,zheng2021global}. 
However, there are few feature-based methods available at the moment. For example, Shi et al.~\cite{shi2020weakly} proposed a novel attention-based method DGAM, an attention-based refinement module. The DGAM~\cite{shi2020weakly} is designed for separating the action instance and non-action instance but overlook the modelling of temporal information. Especially, as shown in~\ref{fig:0}, it is difficult to locate action instances for sport cameras due to complex background information and lots of action subjects~\cite{LIU2022387,gao2022pair}, but the frame-level feature, which can focus on the area of action subject, are promising in improving the performance.

To address this issue, we first propose a novel generative attention mechanism to simultaneously model the frame-level and segment-level video features. Specifically, by using the significant differences of actions' foreground(\textit{i.e.}, action instance) and background(\textit{i.e.}, non-action instance), our model is trained to represent the attention of the frame-level and segment-level features. Modeling the frame-level feature can focus on the action subjects of the videos, while modeling the segment-level features is effective for learning the adaptive relationships, which both can learn a feature extraction model with strong abilities in action instance modeling. On the whole, we introduce a generative attention model-based feature semantics (GAF), wherein we leverage GAF derived from frame-level features (within each feature) and GAF derived from segment-level features (across features) to enhance the performance of the backbone subnet. This amalgamation yields GAF based on 3D CNNs with both frame-level-GAF and segment-level-GAF, tailored for the analysis of video action.
% In the meanwhile, considering that most TAD methods mainly focus on the optimization of region proposal and overlook the feature level optimization, we combine the action detector and generative attention model(feature level) to model the original feature semantics.

To our best knowledge, we are the first to model the feature semantic information and study the differences of actions' foreground and background in some video action analysis tasks(\textit{e.g.}, video action recognition, video action detection), and design a GAF that prominently boosts action-related task performance. Our main contributions are summarized as follows:

\begin{itemize}
  
\item
To leverage the feature semantic information of videos, we propose a novel relation attention module designed to effectively harness both frame-level and segment-level features. By leveraging temporal feature semantic information, this module facilitates the modeling of attention mechanisms, thereby enhancing the extraction of temporal sequences and the modeling of video actions.

% extract the temporal sequence and locate the action boundaries.

\item
To capture the frame-level features, we construct a generative attention model aimed at discerning the semantic disparities in features between foreground and background elements for each action objective of the videos.

\item
Based on extensive experiments, we demonstrate that our GAF yields remarkable performance gains on two widely used video action detection datasets, outperforming other attention-based methodologies. Furthermore, our approach validates the efficacy of modeling action features across additional tasks(\textit{i.e.}, video action recognition), where it also exhibits significant improvements.
% On THUMOS14, our GAF achieves an average mAP of 52.7\% when IOU is from 0.3 to 0.7, the new state-of-the-art(SotA) of the CNN-based methods for TAD tasks. On ActivityNet v1.3, our method also verifies the effectiveness of modeling action feature relationships. Especially, for complex background and small objective action detection tasks, our method achieves better performance(around 3.6\% improvement compared with the AFSD~\cite{lin2021learning} on THUMOS14).

\end{itemize}

The remainder of the paper is organized as follows. In Section~\ref{sec: Related work}, We discuss related studies on temporal action detection, attention module and generative model respectively. In Section~\ref{sec:Method}, we present the details of our model's overview and theoretical basis. In Section~\ref{sec:Experiment}, we report the implementation details, evaluation results and ablation study. In Section~\ref{sec:Conclusion}, we conclude the paper.

\section{Related Work}\label{sec: Related work}

Related work on video action analysis, generative model and attention module are briefly reviewed in this section.

\subsection{Video Action Analysis}
Early works use hand-craft methods to extract features and generate region proposals by sliding windows~\cite{yuan2016temporal}. Recently, with the development of CNN~\cite{guo2018double,guo2019nat,cao2019multi} and Transformer~\cite{liu2021end} in image and video domains, video action analysis tasks achieve great progress. 
Generally, Video Action Analysis research has primarily emphasized two core tasks: video action recognition~\cite{kong2022human,zhou2023learning} and video action detection~\cite{shi2023tridet,wang2023weakly}. Concurrently, researchers have delved into innovative methodologies including graph-based representations~\cite{zhao2023re2tal}, attention mechanisms~\cite{sridhar2021class}, and generative models~\cite{wang2023weakly} to augment the efficacy of Video Action Analysis systems.
% Recently, inspired by the success of object detection in the image analysis domain, many works follow the paradigms of object detection, because both of them try to find meaningful regions and then recognize them. Specifically, there are mainly two categories: two-stage frameworks~\cite{chao2018rethinking,dai2017temporal,heilbron2017scc} and one-stage frameworks~\cite{buch2019end,lin2021learning,ning2021srf,yang2020revisiting}. 
However, these methods mainly rely on the quality of meaningful regions and the precision of classifiers, so most of them only focus on the optimization of action region proposals. Diverging from these conventional methods, our work places particular emphasis on feature semantics, recognizing its crucial role in enhancing the performance of temporal action modeling tasks.

\subsection{Generative Model}
The generative model is developing rapidly in recent years~\cite{goodfellow2014generative,kingma2013auto}.
GAN~\cite{goodfellow2014generative} is aiming at maximizing the approximate real data distribution by reducing the error between the 
generated variables and the output of a neural network. This approach, however, results in a decrease in the sample diversity and 
training instability.
VAE~\cite{kingma2013auto} optimizes the evidence lower bound(ELBO) to reduce the reconstruction error, which is aiming at 
approximating the real data distribution. Specifically, for the VAE, the target of optimization is that for a 
given distribution, the data distribution can be represented and approximated by sampling latent variables. However, VAE 
isn't suitable for CNN to model a distribution with multiple modes~\cite{sohn2015learning}. In our work, in consideration of TAD tasks, we 
use conditional variational autoencoder(CVAE) generated by the probability variational model, exploiting CVAE~\cite{sohn2015learning} to model
the distribution of feature, which is conditioned on attention-based value.

\subsection{Attention Module}
Attention-based methods have been applied in many fields. \cite{bahdanau2014neural} presents the attention mechanism,
which can effectively search for parts of the source sentence that are relevant to the prediction of a target word in the NLP. Then, based on attention mechanisms, \cite{vaswani2017attention} introduces Transformer and dispenses 
with recurrence and convolutions. Due to the great success of attention modules, in the field of video action analysis, many studies 
~\cite{chen2019relation,li2020graph,sridhar2021class} start to focus on the use of attention mechanisms. Most of these methods~\cite{li2020graph,chen2019relation}, however, only focus on the enhancement of the attention mechanism in the action region but ignore the relation inside each feature.
A more closely related generative attention model is DGAM~\cite{shi2020weakly}. DGAM is designed to model weakly supervised learning, aiming at directly generating coarse temporal action intervals with the probabilistic model using context-action differences. Unlike DGAM, our model employs temporal information which focuses on precisely localizing temporal action intervals using action instances accurately identified within the interval through foreground-background separation. Especially, the GAF proposed by us leverages the significant feature semantics between the foreground and background of action instances to locate the action boundaries and classify the action categories for various action analysis tasks.

\section{Method}\label{sec:Method}
In this section, we first introduce our GAF-based video action analysis framework and then elaborate on the corresponding technical details.

\begin{figure*}[h]
  \centering
  \includegraphics[width=0.95\textwidth]{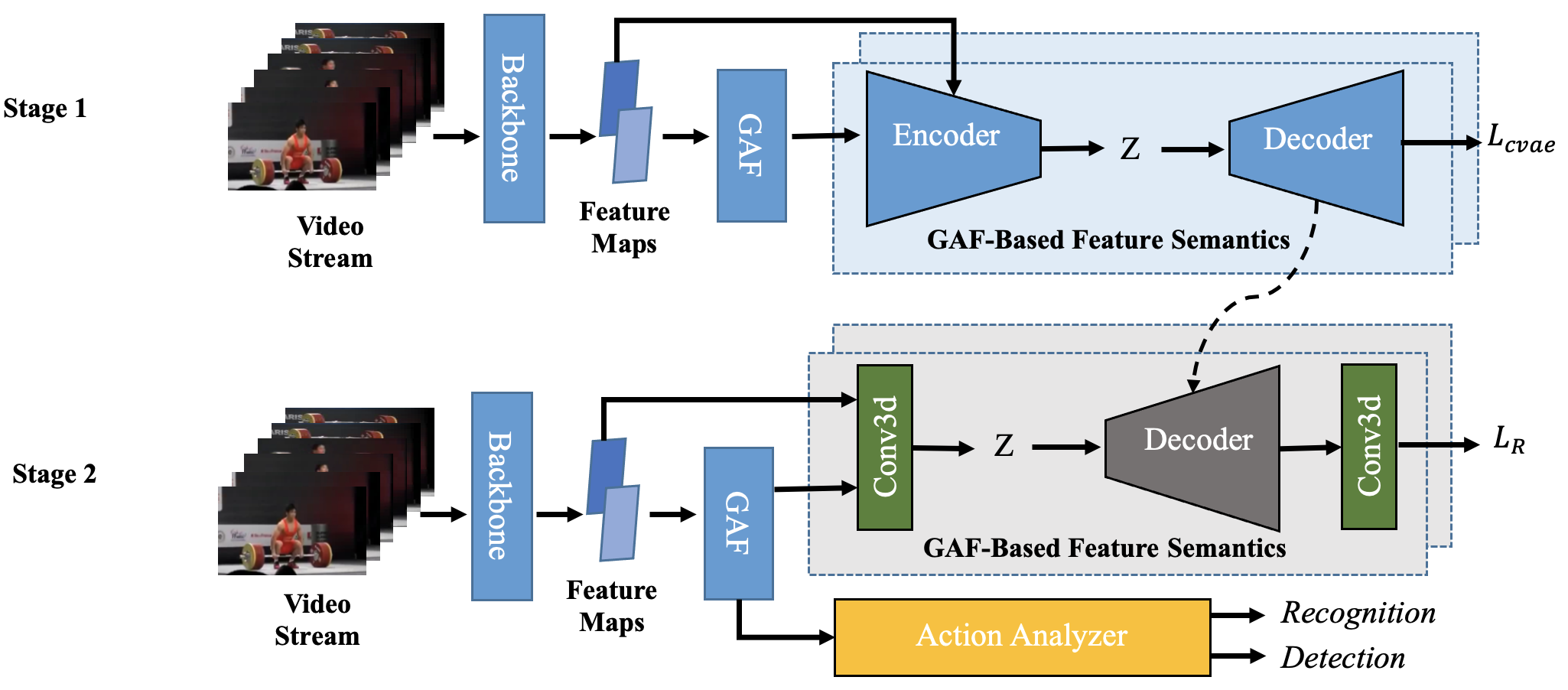}
  \caption{The layout of a generic video action analysis architecture enhanced by our GAF attention mechanism, where frame-level attention feature and segment-level attention feature are trained alternatively in two steps. In step 1, the frame-level-GAF is trained with reconstruction loss $ L_{cvae} $ to learn the frame-level feature relations. In step 2, the segment-level-GAF is updated with representation loss $ L_R $, which learns the adaptive dependencies among the features and fuses frame- and segment-level relations.}
  \label{fig:1}
\end{figure*}

\subsection{Overview}
We first describe the design of a generic video action analysis architecture(Fig.~\ref{fig:1}), which consists of a pre-trained backbone  that extracts feature semantics and a video action analyser that is trained with feature semantic fused attention as input to finish video action analysis tasks. The framework of our model is shown in Fig.~\ref{fig:1}. The input of the feature extraction model is a 
clip of video frames, which forwards to the backbone. Then, through the backbone, our model extracts the basic feature map:
\begin{align}
  F = \left\{ f^i \in B\times C \times H\times W, i = 1, \cdots , L \right\},  
\end{align}
where B, C, H, W and L represent the batch size, channels, height, width and temporal length of the feature maps respectively. For basic feature maps, we utilize our 
attention-based mechanism GAF to enhance action features, which forms semantics-enhanced action feature $ F^{\prime}$.
Finally, we use an anchor-free detector based on FPN~\cite{lin2021learning} to detect action instances for video action detection task, and use a classifier for video action recognition task. 

\begin{figure}[t]
 \centering
 \setlength{\belowcaptionskip}{-0.4cm}
  \subfloat[Frame-level-GAF]{
      \label{Fig2.a}
      \includegraphics[width=0.23\textwidth]{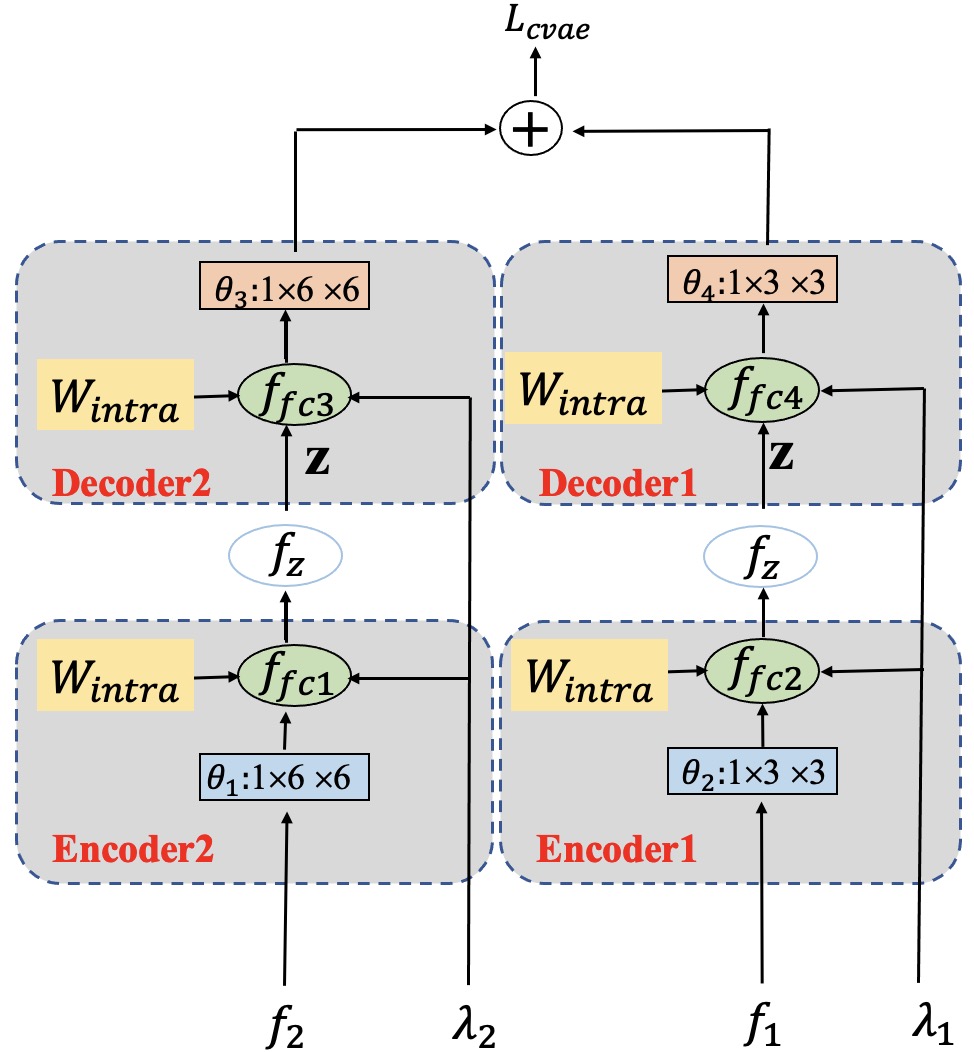}
  }
  \subfloat[Segment-level-GAF]{
      \label{Fig2.b}
      \includegraphics[width=0.23\textwidth]{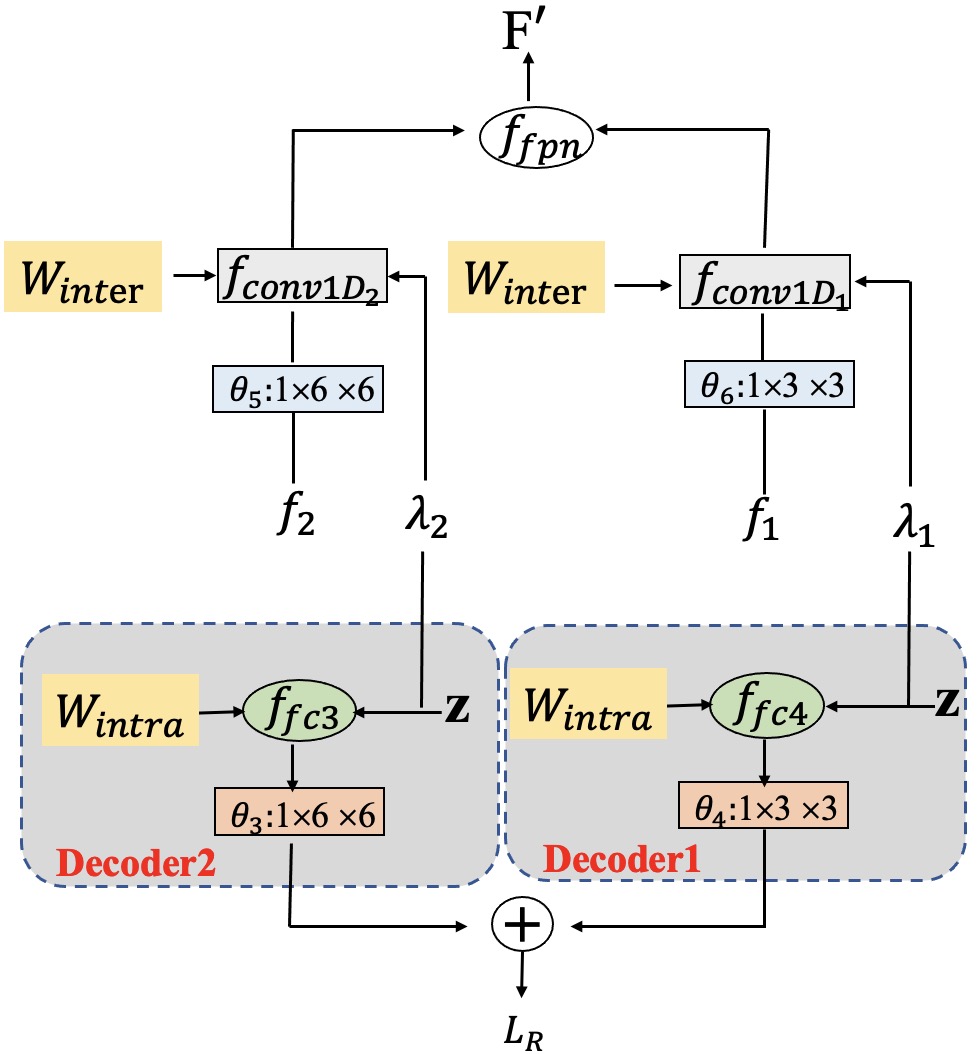}
  }
  \caption{\textcolor{blue}{The architecture of our Generative Attention Model-Based Feature Semantics(GAF). It includes frame-level attention-based GAF and segment-level attention-based GAF.}}
  \label{fig:2}
\end{figure}

Our GAF module enhances feature semantics by modeling frame- and segment-relations of features. In specific, our GAF includes frame- and segment-attention modules based on feature semantics, which learn the corresponding frame-relations of the action feature semantic information 
and segment-relations of the action temporal semantic information respectively. Specifically, as shown in Fig.~\ref{fig:1}, frame-GAF explores the relation of action instances inside each feature, which is trained with $ L_{cvae} $ to reconstruct the representation of 
feature semantics. The segment-GAF learns the dependencies between different features, which is trained with $ L_{clf} $ and $ L_{reg} $ to accomplish activity classification and boundary prediction respectively. We will introduce more details about frame-level attention module and segment-level attention module in a later section.

\subsection{Generative Attention Model-Based Feature Semantics}
In practice, our method learns the frame- and segment-relations of the temporal action features, which is aiming at enhancing the action extraction capabilities of the video action analysis framework. As Fig.~\ref{fig:2} shows, our attention model consists of two submodules, frame- and segment-GAFs. In the following paragraphs, we first introduce the definition of the attention-based video action analysis problem.

In attention-based video action analysis, the target aims at generating attention value to pay more attention to the region of action subjects. 
However, for the generative-based attention model, the target is aiming at generating the attention value $ \lambda $ by leveraging temporal action regions and action classes. So, the optimization target of our model is: 
\begin{align}
  \max \limits_{\lambda \in \left[0,1\right]} \log p \left(\lambda \rvert F, C, I\right),
\end{align}
where $ \lambda \in \left[0,1\right] $ is the attention of feature $ F $, $ C $ is the label of action class, $ I $ is the label of action boundaries~\footnote{The variable "$ I $``  is solely pertinent to the video action detection task, whereas it holds no relevance to the video action recognition task. Importantly, its absence does not impact the derivation of the formula.}. We utilize Bayes' theorem to simplify the optimization problem and get the result:

\begin{align}
\max \limits_{\lambda \in \left[0,1\right]} \log p \left(C\rvert F, \lambda \right) + \log p \left(I\rvert F, \lambda \right) + \log p \left(F\rvert \lambda \right) + \log p \left(\lambda \right).
\end{align}

Specially, we discard the constant term $ \log p \left( F, C, I\right) $ and set $ \lambda $ as a uniform distribution, so our optimization target is transformed to 
\begin{align}
\max \limits_{\lambda \in \left[0,1\right]} \log p \left(C\rvert F, \lambda \right) + \log p \left(I\rvert F, \lambda \right)
+ \log p \left(F\rvert \lambda \right).
\end{align}

As far as concerned of this formulation, the first two terms $ \log p \left(C\rvert F, \lambda \right)$ and
$ \log p \left(I\rvert F, \lambda \right)$, which correspond with segment-attention model(Fig.~\ref{Fig2.b}) of our GAF, prefer $ \lambda $ 
with segment-dependencies for video action analysis, which leverages temporal action information to represent the segment-relations among features. In the 
meantime, the third term $ \log p \left(F\rvert \lambda \right) $, which corresponds with frame-attention model(Fig.~\ref{Fig2.a}) of our 
GAF, learns to represent features correctly predicted from the attention $ \lambda $, which leverages feature semantic
information to represent the frame-relations inside each feature. Due to the feature semantic difference between the action instance 
and non-action instance, our GAF module encourages the model to enhance the attention value of the action instance and restrain the attention 
value of the non-action instance. Inspired by the success of the generative model, we leverage CVAE to represent $ \log p \left(F\rvert \lambda \right) $, 
and reconstruct the feature $ F $ accurately, which is designed to make a model concentrated on the action subject for the  video action analysis tasks.

\subsubsection{Frame-level Attention}
In our frame-level attention approach, we specifically target the regions corresponding to action instances within each frame of action features, particularly amidst complex background. Considering the feature semantic difference between foreground and background, we utilize the CVAE~\cite{sohn2015learning,shi2020weakly} to model the feature distribution(\textit{i.e.}, action instance and non-action instance). As similar with VAE, CVAE introduces a latent variable $ Z_t $, then attempts to reconstruct feature variable $ f_t $ from latent variable $ Z_t $ and conditional variable $ \lambda_t $:
\begin{align}
  p_\psi \left(f_t \rvert \lambda_t \right) = \mathbb{E}_{p_\psi\left( z_t\rvert \lambda_t \right)}\left[ p_\psi\left( f_t\rvert \lambda_t, Z_t\right)\right],
\end{align}
where $ p_\psi $ represents the CVAE model, $ p_\psi\left( z_t\rvert \lambda_t \right) $ indicates the prior model, and
$ p_\psi\left( f_t\rvert \lambda_t, Z_t\right) $ is the decoder procedure of the generative model, which is a conditional distribution.

During the training of the Frame-GAF, we approximate the Encoder procedure $ p_\psi \left(z_t \rvert f_t, \lambda_t \right) $ of generative model by the Gaussian distribution,
\begin{align}
   q_\phi \left(z_t \rvert f_t, \lambda_t \right) = \mathcal{N}\left(z_t \rvert \mu_\phi,\Sigma_\phi \right),
\end{align}
where $ \mu_\phi $ and $ \Sigma_\phi $ represent the encoder parameters of generative model. We minimize the evidence lower bound (ELBO) loss of CVAE $ \mathcal{L}_{CVAE} $:
\begin{equation}
  \begin{split}
    \mathcal{L}_{CVAE} &= - \frac{1}{L}\sum_{i=1}^L \log p_\psi\left(f_t\rvert \lambda_t, z_t^{\left(l\right)}\right)\\
    &\ \ \ + \beta \cdot KL\left( q_\phi\left(z_t\rvert f_t, \lambda_t\right) \parallel p_\psi\left(z_t\rvert \lambda_t\right)\right),
  \end{split}
\end{equation}
where $ z_t^{\left(l\right)} $ is generated by $ rvert x_t $ and $\lambda_t $, which is the Encoder of generative model,  and $ \beta $ is a hyper-parameter.

In our framework, the $ \lambda_t $ isn't only generated by training a neural network but optimized by a generative model. Therefore,
we propose to train Frame-GAF and Segment-GAF simultaneously. Specifically, we first update Frame-GAF with initial value $ \lambda_t $
, which is given by the Segment-GAF, and then train Segment-GAF with the fixed Frame-GAF. We trained GAF in an alternating way. In our model, as shown 
in Fig.~\ref{Fig2.a}, we use $ x_i $ and $ \lambda_i $ as the input, and $ W_{intra} $ as the learnable network weight, to generate 
reconstructed feature map $ \widehat{F} $, which is aiming at optimizing loss function $ \textit{l}_{cvae} $. In specific, firstly, we use \emph{Encoder} to generate the 
intermediate variable $ z $, which leverages function $ f_z $ to generate latent variable $ Z $ through feature $ F $ and attention 
$ \lambda $, as follows:
\begin{align}
  Z = f_z\left(f_{fc_i}\left(\lambda_i, \theta_i\left(f_i\right)\right)\right),
\end{align}
where $ f_{fc_i} $ is a liner model, and $ \theta_i $ is a $ 1\times1\times1 $ 3D convolution, which is aiming at 
concatenating $ \lambda_i $ for $ f_{z} $ by reducing the channels' number and adding learnable model parameters. $ f_z $ is a mean and variance function for the intermediate variable $ z $, which generates the latent variable $ Z $. Then, we use \emph{Decoder} to generate the reconstructed feature map $ \widehat{F} $,  which leverages latent variable $ Z $ and attention 
$ \lambda $ , as follows:
\begin{align}
  \widehat{F} = \theta_i^{\prime}\left(f_{fc_i}\left(Z,\lambda_i\right)\right),
\end{align}
where $ \theta_i^{\prime} $ is a $ 1\times n \times n $( $ n $ is the kernel size of input feature) 3D deconvolution, which is aiming at generating the reconstructed feature map $ \widehat{F} $ by adding the number of channels. We use $ \textit{l}_{cvae} $ to measure the difference between $ \widehat{F} $ and $ F $.

\subsubsection{Segment-level Attention and Attention Fusion}
For the Segment-attention module, our goal is to acquire insights into the relationships between action instances across different features. As shown in Fig.~\ref{Fig2.b}, we use the enhanced feature $ F^{\prime} $ as the input of action detector, as follows:
\begin{align} 
  F^{\prime} = f_{fpn}\left(f_{conv1D_i}\left(\lambda_i, \theta_i\left(x_i\right)\right)\right),
\end{align}
where $ \theta_i $ is a $ 1\times n \times n $ ($ n $ is the kernel size of input) 3D deconvolution, which is aiming at fusing $ \lambda $ with $ F $, $ f_{fpn} $ is the function of FPN layer, which averages two layers as the input of Detector. During the training of Segment-GAF and Action Analyzer, the Segment-GAF is frozen. In the meanwhile, the Frame-GAF module and the Action Analyzer module are updated.

The Segment-GAF learns the segment-relations between features for optimizing video action related tasks. 
Specifically, we take advantage of the attention value $ \lambda $ as a weight to perform average pooling on the features of the video
and generate the \emph{action-instance} feature $ f_{ai} \in R^d $ given by
\begin{align}
  f_{ai }= \frac{\sum_{t=1}^T \lambda_t f_t}{\sum_{t=1}^T \lambda_t}.
\end{align}
Similarly, we utilize $ 1-\lambda $ to represent the \emph{non-action instance} feature $ f_{n-ai} $~\cite{nguyen2019weakly}.

To optimize $ \lambda $, we encourage higher capability of the action-instance feature $ f_{ai} $ and contemporaneously restrain 
capability of the non-action instance features $ f_{n-ai} $, which is aiming at learning temporal information of feature semantics.
This amounts to minimizing the following loss~\footnote{The video action recognition task exclusively comprises the component $L_{clf}$.}:
\begin{align}
  \begin{split}
  & L_{clf} = L_{clf_{ai}} + L_{clf_{n-ai}} \\ 
  & = - \log p_\varphi \left(c\rvert f_{ai}\right) -  \log p_\varphi \left(0\rvert f_{n-ai}\right),\\
  \end{split}\\
  \begin{split}
  & L_{reg} = L_{reg_{ai}} + L_{reg_{n-ai}} \\
  & = - \log p_\varphi \left(I\rvert f_{ai}\right) -  \log p_\varphi \left(\bar{I}\rvert f_{n-ai}\right),\\  
  \end{split}\\
  & L_d = \alpha L_{clf} + \beta L_{reg},
\end{align}
where $ \alpha $ and $ \beta $ are hyper-parameters, $ p_\varphi $ is an action detector module, $ c $ is the label of categories, $ I $ is the label of boundaries.

During the training, as shown in Fig.~\ref{Fig2.b}, we use \emph{Decoder} to generate reconstructed feature map $ \widehat{F} $, which fuses feature map $ F $ with attention $ \lambda $, and minimizes the reconstruction loss $ L_R $:
\begin{align}
\label{con:r}
\begin{split}
  \mathcal{L}_{R} &= -\sum_{t = 1}^T \log \left\{p_\psi \left(f_t \rvert \lambda_t \right)\right\} \\ 
  & \backsimeq -\sum_{t=1}^T \log \left\{\frac{1}{L}\sum_{l=1}^L p_\psi\left(f_t\rvert\lambda_t,z_t^{\left(l\right)}\right)\right\} \\
  & \backsimeq \sum_{t=1}^T \parallel f_t - f_\psi\left(\lambda_t, z_t\right)\parallel^2,
\end{split}
  \end{align}
where $ z_t^{\left(l\right)} $ is generated by $z_t$ and $ \lambda_t $ in the encoder of generative model. Especially, following~\cite{shi2020weakly}, in the last step, we set $ L $ as 1.

As shown in Fig.~\ref{Fig2.b}, we use $ L_R $ loss to fuse segment-level attention between features with frame-level attention inside features as follows:
\begin{align}
  X_R=\theta_i^{\prime}\left(f_{fc_i}\left(Z,F\left(\lambda_i\right)\right)\right),
\end{align}
where $ F\left(\lambda_i\right) $ is the feature map enhanced by segment-GAF, $ Z $ is the information including frame-relations through 
frame-GAF. We will discuss the effectiveness of each design in the next section.

\section{Experiments}\label{sec:Experiment}
In this section, we first introduce datasets and evaluation metrics. Then, we analyze our model's effectiveness through the main result 
and ablation study.

\subsection{Datasets and Evaluation Metrics}
To validate the effectiveness of our GAF, three datasets are employed to evaluate our proposed generative attention-based video action analysis method: UCF101~\cite{soomro2012ucf101} for video action recognition, THUMOS14~\cite{THUMOS14} and ActivityNet v1.3~\cite{caba2015activitynet} for video action detection. 

\textbf{UCF101} comprises 13,320 action videos across 101 action categories, which is partitioned into three official splits, with each split dividing the dataset into training and test sets at a ratio of 7:3.

\textbf{THUMOS14~\cite{THUMOS14}} contains 20 categories of activities and about 413 untrimmed videos. The dataset is composed of two 
parts: the validation set and the testing set, which contain 200 and 213 videos, respectively. Whereas, compared with the ActivityNet v1.3 
dataset, the THUMOS14 dataset is more difficult because each video in THUMOS14 contains a large amount of non-action instance 
fragments and multiple action instances categories~\cite{chen2019relation}.

\textbf{ActivityNet v1.3~\cite{caba2015activitynet}} contains 200 categories of activities and around 20000 videos in total. The 
dataset is divided into three parts: training, validation, and test sets, which contain 10024, 4926 and 5044 videos respectively.

\textbf{Evaluation Metrics}. Adhering to the established evaluation protocol, for video action detection datasets, we reported the mean Average Precision(mAP) at the different overlaps intersection over union(IoU). The mAP results were calculated through the ActivityNet official codebase\footnote{https://github.com/activitynet/ActivityNet/tree/master/Evaluation} code. Conversely, for video action recognition datasets, we report both Top-1 and Top-5 accuracy metrics to assess the performance.

\setlength{\tabcolsep}{4pt}
\begin{table*}[h]
\begin{center}
\caption{
% Performance comparison with state-of-the-art methods on THUMOS14 and ActivityNet v1.3, measured by mAP at different IoU thresholds on THUMOS14 and ActivityNet v1.3.
Performance comparison with video action detection methods on THUMOS14 and ActivityNet v1.3, measured by mAP at different IoU thresholds. For fairness, all of attention-based method use the same feature extraction backbone.
}
\label{table:1}
\setlength{\tabcolsep}{3.5mm}{
\begin{tabular}{c c c c c c c c c c c }
\hline\noalign{\smallskip}
\multicolumn{1}{c}{\multirow{2}*{Type}} & \multirow{2}*{Model} & \multirow{2}*{Attention} & \multicolumn{5}{c}{THUMOS14} & \multicolumn{3}{c}{ActivityNet v1.3} \\
\cline{4-11}
\multicolumn{1}{c}{}& & &0.1& 0.2 &0.3 & 0.4 & 0.5  & 0.5 & 0.75 & 0.95 \\
\noalign{\smallskip}
\hline
\noalign{\smallskip}
\multirow{8}*{Non-Attention} & BSN~\cite{lin2018bsn} & - & - & - & 53.5 & 45.0 & 36.9  & 46.5 & 30.0 & 8.0 \\
\cline{4-11}
\multirow{8}*{}& BU-TAL~\cite{zhao2020bottom} &- & - & - & 53.9 & 50.7 & 45.4  & 43.5 & 33.9 & 9.2\\
\cline{4-11}
\multirow{8}*{}& GTAN~\cite{long2019gaussian} &- & 69.1&63.7 & 57.8 & 47.2 & 38.8 & 52.6 & 34.1 & 8.9 \\
\cline{4-11}
\multirow{8}*{}& DBS~\cite{gao2019video} & -&56.7& 54.7 & 50.6 & 43.1 & 34.3  & 43.2 & 25.8 & 6.1 \\
\cline{4-11}
\multirow{8}*{}& G-TAD~\cite{xu2020g} & - &-& - & 54.5 & 47.6 & 40.2  & 50.4 & 34.6 & 9.0 \\
\cline{4-11}
\multirow{8}*{}& GCM~\cite{zeng2021graph} & - &72.5& 70.9 & 66.5 & 60.8 & 51.9  & 51.0 & 35.2 & 7.4 \\
\cline{4-11}
\multirow{8}*{}& ContextLoc~\cite{zhu2021enriching} & - &-& - & 68.3 & \textbf{63.8} & 54.3 & \textbf{56.0} & 35.2 & 3.6 \\
\cline{4-11}
\multirow{8}*{}& AFSD~\cite{lin2021learning} & - &72.2& 70.8 & 67.1 & 62.2 & 55.5 & 52.4 & 35.3 & 6.5 \\
\cline{4-11}
% \multirow{8}*{}& PcmNet ~\cite{Qin202248} & Neurocomputing2022 &-& - & 61.5 & 55.4 & 47.2 &51.4 &36.1 &\textbf{9.5}\\
\hline
\multirow{6}*{Attention-Based}&AGCN~\cite{li2020graph} & Graph &59.3& 59.6 & 57.1 & 51.6 & 38.6  & 30.4 & - & - \\
\cline{4-11}  

\cline{4-11}  
\multirow{6}*{}& RAM~\cite{chen2019relation} & Relation &65.4& 63.1 & 58.8 & 52.7 & 43.7 & 37.0 & 23.1 & 3.3 \\
\cline{4-11}
\multirow{6}*{}& TadTR~\cite{liu2021end} & Transformer &72.7 & 69.9 &62.4 &57.4 &49.2 &- &- &- \\ 
\cline{4-11} 
\multirow{6}*{}& CSA~\cite{sridhar2021class} & Class &-& - & 64.4 & 58.0 & 49.2 & 51.9 & \textbf{36.9} & 8.7 \\ 
\cline{4-11} 
\multirow{6}*{}& Ours & Frame-segment &\textbf{73.6}& \textbf{72.1}& \textbf{68.9} & 62.1 & \textbf{55.9}  & 53.4 &\textbf{36.9} &5.9 \\
\hline
\end{tabular}}
\end{center}
\end{table*}
\setlength{\tabcolsep}{4pt}

\begin{table}[tbp]\centering
\caption{Performance comparison with video action recognition methods on UCF101.}
\label{tab4}
    \setlength{\tabcolsep}{3mm}{
    \begin{tabular}{cccccccc}
    \hline
     Model   &Top-1   &   Top-5 & Model    &Top-1   &   Top-5\\
     \hline\noalign{\smallskip}
     I3D  &91.9    &   98.8 & Top-I3D &64.1    &   82.1\\
     CSA  &92.0    &   99.0 & CSA &64.3    &   82.6\\
     Ours  &\textbf{92.4 }   &   \textbf{99.1} & Ours &\textbf{64.7}    &   \textbf{82.9} \\
     \hline
\end{tabular}}

\end{table}

\subsection{Implementation Details}
For the THUMOS14 training process, we first sampled both RGB and optical flow data at 25 fps for each video. We set 
each video clip length $ \emph{T} = 256 $. For the training process on ActivityNet v1.3, we sampled frames using different fps 
and ensured the length of each video was  $ 768 $. On the THUMOS14 and ActivityNet v1.3 datasets, both the frame height and weight 
size were set to $ 96 $. For data augmentation in training, we used random crop and horizontal flipping.
On UCF101, we use tvl1~\cite{brox2009large} to extract optical frames. The length of the clip is set to 64. We resize the frame to 256 for UCF101.

Our model was trained for 20 epochs, which set the training video batch size to 1, applying the Adam~\cite{kingma2014adam} optimizer with a learning rate of $ 10^{-5} $, and a weight decay of $ 10^{-3} $ . During the training, we trained Segment-GAF and Frame-GAF alternatively, \textit{i.e.}, as the stage 1 in Fig.~\ref{fig:1} shows, we first updated Frame-GAF with initial label of $ \lambda_t $ given by the Segment-GAF, and then, as the stage 2 in Fig.~\ref{fig:1} shows, we trained the Segment-GAF with the fixed Frame-GAF. 
During the test phase, the RGB and optical flow results are averaged to give the final location and class score.

\subsection{Main Results}
In Table~\ref{table:1}, We compare our GAF with current video action detection methods on Thumos14. On THUMOS14, our model outperforms the 
current attention-based competitor AGCN, CSA and RAM, especially 6.7\% on mAP@0.5 with CSA. The obvious improvement is proved to be 
effective, making the model more practical in real video action detection tasks. Note that CSA also has an attention module, but its overall performance 
of the model is worse than ours, proving the superiority of our attention-based architecture. On ActivityNet v1.3, our result is also far
better than most of the methods, where mAP is comparable to the other results. 
% It is noteworthy that all of attention-based methods have less mAP than ours on THUMOS14 and most of attention-based methods have less mAP than ours on 
% ActivityNet v1.3.

\begin{figure}[t]
  \centering
  \includegraphics[width=0.45\textwidth]{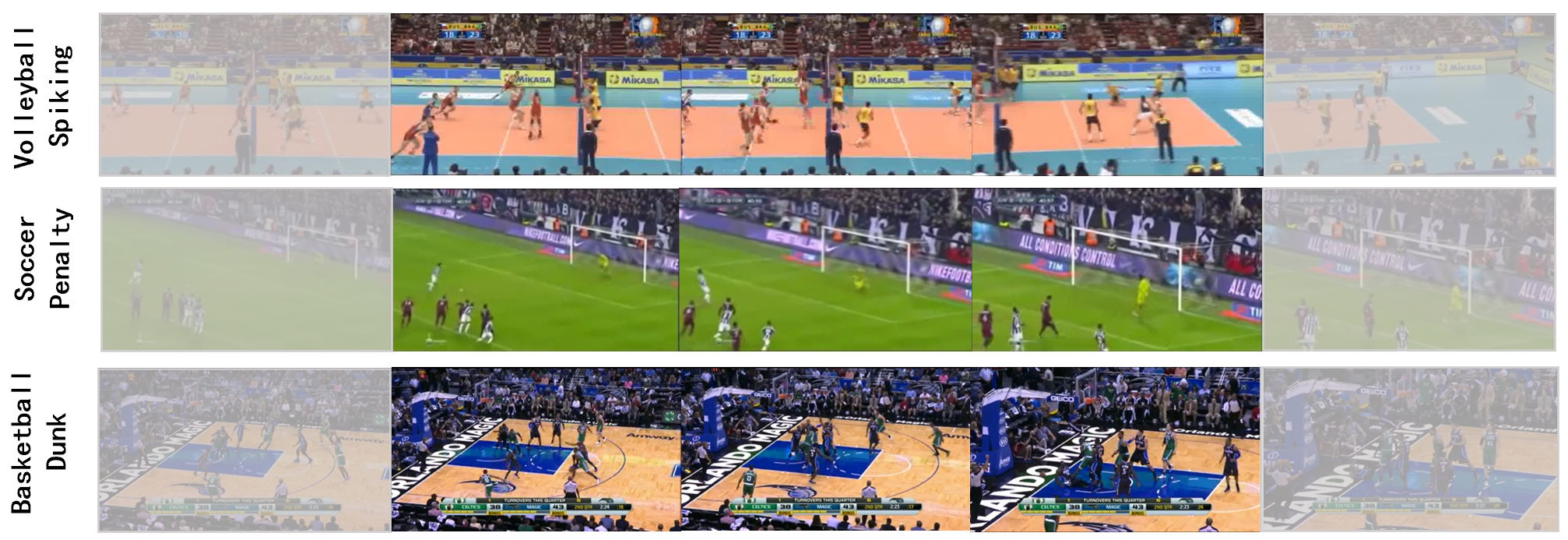}
  \caption{Examples of action instances predicted by THUMOS14. Each row shows frames sampled within or outside the temporal range of 
  the detected action. Faded frames indicate locations outside the detection range, indicating localization ability.}
  \label{fig:3}
\end{figure}

For a thorough comparison, we conduct existing action detection methods into two categories: Non-Attention and Attention-Based methods. Most of 
Non-attention methods explore effective solutions in optimizing action region, including anchor-based methods~\cite{lin2021learning,long2019gaussian,zeng2021graph,zhu2021enriching} and 
actionness-based methods~\cite{lin2018bsn,zhao2020bottom,gao2019video,xu2020g}. Among these methods, GCM~\cite{zeng2021graph} shows strong 
performance under the threshold of 0.1 and 0.2. The reason is that GCM utilized the graph convolutional module to represent the relation
of action instances, which is effective for localizing more action instances. Different from GCM(and other methods), we not only consider the 
relation of action instances by using feature semantics, but also leverage significant differences between the foreground and background 
of action instances, so we achieve better performance under the threshold from 0.1 to 0.5. Among these methods, Attention-Based methods 
achieve better performances compared with their baseline. To be precise, these methods focus on the optimization of the proposal region. 
In contrast, we use GAF that effectively explores the inter and intra-relations of features and leverages the temporal feature semantics 
information to model attention mechanism, which achieves better performance compared with baseline AFSD~\cite{lin2021learning}(up to 
1.8\% improvement on THUMOS14 at IoU = 0.3) .

To evaluate the performance of our GAF, we conduct experiments on the ActivityNet v1.3 dataset, the largest temporal action detection 
dataset, to assess the effectiveness of our GAF. As shown in~\ref{table:1}, our model also has the comparable performances with CSA under all IoU thresholds and average mAP.

Simultaneously, to demonstrate the efficacy and advancement of our method, we undertake experimental analyses within the broader domain of video action recognition tasks. As depicted in Table~\ref{tab4}, we incorporate some methods into two prevalent backbone architectures(\textit{i.e.},I3D~\cite{carreira2017quo}, Top-I3D~\cite{wang2023generative}) to substantiate their performance in comparison. In light of the absence of a dedicated two-stream action recognition attention algorithm, we opt to reproduce the most effective class-semantic attention algorithm from the realm of action detection, for the purpose of comparison. As shown in Table~\ref{tab4}, compared to the I3D, our module contributes an increase of $0.5\%$, $0.3\%$ and the performance finally reaches $92.4\%$, $99.1\%$ on the Top1 and the Top5. Compared to the Top-I3D model, our module contributes an increase of $0.6\%$, $0.8\%$ and the performance finally reaches $64.7\%$, $82.9\%$ on the Top1 and the Top-5. Both sets of results exhibit superiority over the class-semantic attention method (CSA~\cite{sridhar2021class}), indicating that the incorporation of our module enables effective representation of feature semantics and leads to improved model performance.

% \begin{table}[t]
%   \begin{center}
%   \caption{
%   % Performance comparison with state-of-the-art methods on THUMOS14 and ActivityNet v1.3, measured by mAP at different IoU thresholds on THUMOS14 and ActivityNet v1.3.
%   Performance comparison with attention-based methods on ActivityNet v1.3.
%   }
%   \label{table:5}
%   \resizebox{\textwidth}{!}{
%   \begin{tabular}{|c|c|c|c|c|}
%   \hline\noalign{\smallskip}
%   \multicolumn{1}{|c|}{Model}&\multicolumn{4}{|c|}{ActivityNet v1.3}  \\
%   \multicolumn{1}{|c|}{}&0.5& 0.75 &0.95 &avg. \\
%   \hline
%   TadTR~\cite{liu2021end} & 47.6 & 31.7 & 8.0 & 29.1 \\
%   AGCN~\cite{li2020graph} &30.4& - & - & - \\
%   RAM~\cite{chen2019relation} &37.0& 23.1 & 3.3& 21.1 \\
%   CSA~\cite{sridhar2021class} &51.9& \textbf{36.9} &\textbf{8.7}& \textbf{32.5}\\
%   Ours &\textbf{53.7}& \textbf{36.9}& 3.9 & 31.6\\
  
%   \hline
%   \end{tabular}}
%   \end{center}
%   \end{table}

\begin{table}[t]
\begin{center}
\caption{Per-class breakdown (AP) on THUMOS14, at IoU = 0.5. }
\label{table:2}
\normalsize
\resizebox{0.45\textwidth}{!}{
\begin{tabular}{l c c  l c c }
\hline\noalign{\smallskip}
\  & AFSD~\cite{lin2021learning} & Ours & & AFSD~\cite{lin2021learning} & Ours \\
\noalign{\smallskip}
\hline
\noalign{\smallskip}
Baseball Pitch &75.9  & 68.2  & Hamm. Throw & 33.9 & {\textbf{36.7}} \\
Basket. Dunk  & 47.8 & {\textbf{50.8}} & High Jump  &89.3  &87.3 \\
Billiards & 91.5 & {\textbf{93.3}} & Javelin Throw &37.9 &29.6 \\
Clean and Jerk & 91.2 & 91.0 & Long Jump &75.2  &70.7 \\
Cliff Diving &21.5 &21.4 & Pole Vault &18.1  &15.1 \\
Cricket Bowl. &78.7  &77.3  & Shotput &31.9  &25.8 \\
Cricket Shot & 74.0 & {\textbf{76.6}}  & Soccer Penalty & 58.2 & {\textbf{66.5}}\\
Diving &78.5  &{\textbf{80.6}}  & Tennis Swing &40.3  &38.5 \\
Frisbee Catch &15.5  &13.6  & Throw Discus & 19.1 & {\textbf{21.2}} \\
Golf Swing &74.2  &70.0  & Volley. Spike & 58.3 & {\textbf{67.6}} \\
\hline
\multicolumn{4}{l}{\textbf{\emph{m}AP}} & \multicolumn{1}{ c }{55.5} & \multicolumn{1}{ c }{\textbf{55.9}}\\
\hline
\end{tabular}}
\end{center}
\end{table}

\subsubsection{Per-class breakdown} The per-class AP split of our model is shown in Table~\ref{table:2} along with a comparison to
the top performer~\cite{lin2021learning} on the THUMOS14 leaderboard. In 8 out of 20 classes, our model performs better 
than~\cite{lin2021learning}. Table~\ref{table:2} demonstrates that AFSD with GAF module performs better on some THUMOS14 categories,
especially, for ``Soccer Penalty" and ``Volley. Spike" and so on, we get up to 9.3\% improvement compared with baseline. 
Fig.~\ref{fig:3} shows examples of our model's predictions, including several from small objective and complex background classes. 
Due to the intra- and inter-relations of feature semantics, the model's ability to capture action subject area on video enables it to 
locate temporal boundaries even when action instance subjects are on complex backgrounds.

\begin{figure*}[t]
  \centering
  \includegraphics[width=0.9\textwidth,height=0.6\textwidth]{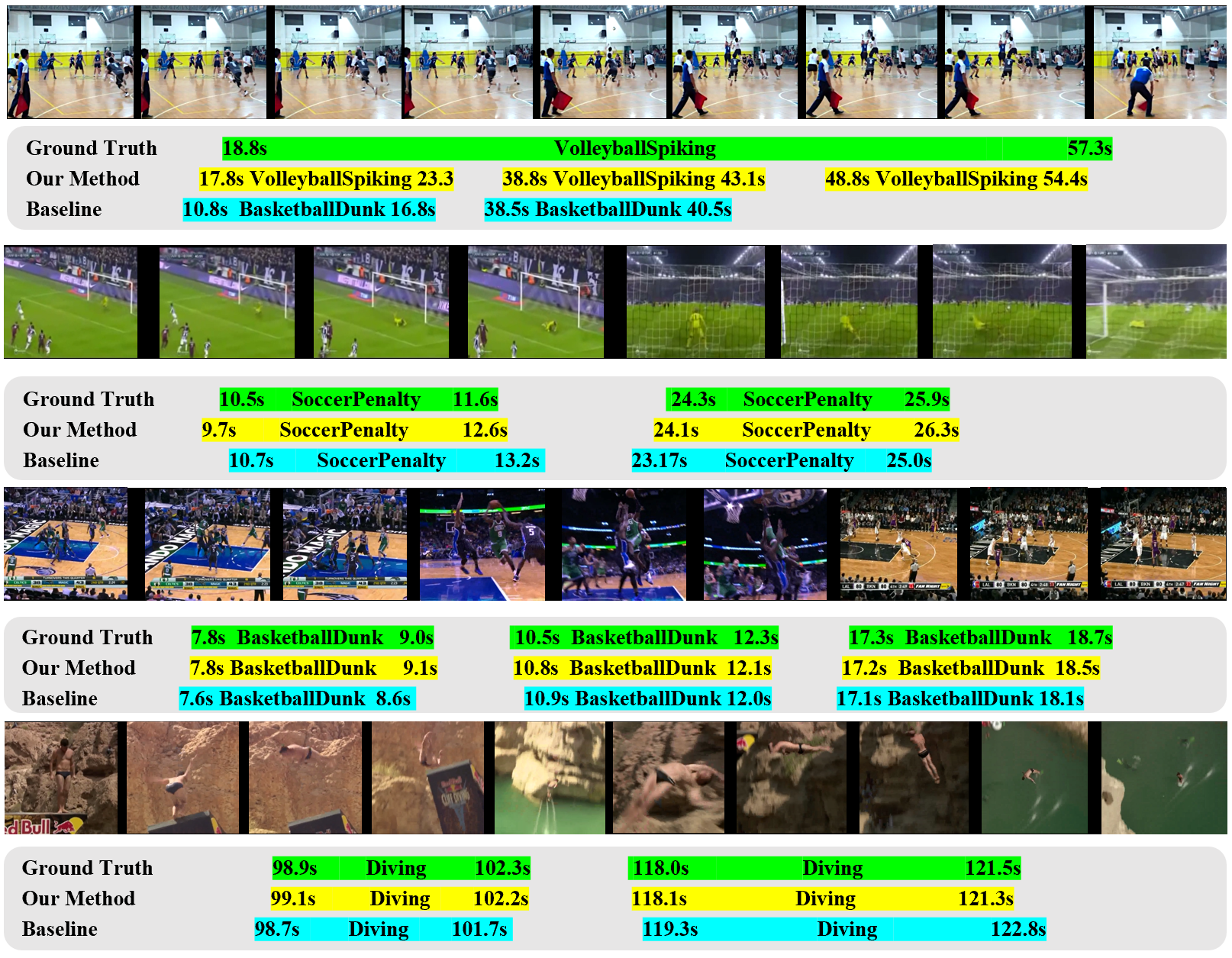}
  \caption{Qualitative results on THUMOS14 dataset. The green segment shows the baseline truth and the yellow segment shows the 
  predicted segment using our model. Our model can accurately localize action instances of various temporal lengths.}
  \label{fig:4}
\end{figure*}

\subsection{Ablation Study}\label{sec:Ablation}

To prove the effectiveness of our GAF, we analyze the impact of each design in this subsection.
\subsubsection{Contribution of each design in GAF}
We study the influence of each component in GAF on overall performance. The $ \mathcal{L}_{ai} $ indicates a basic model, which isn't
equipped with the attention model. Based on $ \mathcal{L}_{ai} $, we added $ \mathcal{L}_{n-ai} $ to distinguish \textbf{action instance} with
\textbf{non-action instance}, which indicates our Inter-attention model of GAF. Then, we added $ \mathcal{L}_R $ to model the intra-relations of
each feature, which indicates our Intra-attention model of GAF.

In Table~\ref{table:3}, we summarize the performance of each design on THUMOS14. Based on Table~\ref{table:3}, our Intra-GAF(only intra-relations inside each feature) brings a performance gain of 2.7\% on IoU = 0.5. The GAF(intra- and inter-relations of features) further contributes an increase of 0.9\% on IoU = 0.5 and the performance of our model finally achieves 55.9\% of IoU = 0.5 threshold.

\begin{table}[t]
  \begin{center}
  \caption{Performance comparison of each design at different overlap IOU on THUMOS14.}
  \label{table:3}
  \normalsize
   \setlength{\tabcolsep}{2mm}{
    \begin{tabular}{c c c c c c c c }
      \hline
      \multicolumn{1}{ c}{ \multirow{2}*{$ \mathcal{L}_{ai} $} }&\multicolumn{1}{c}{ \multirow{2}*{$ \mathcal{L}_{n-ai}$} } &\multicolumn{1}{c }{ \multirow{2}*{$ \mathcal{L}_R $} }& \multicolumn{5}{ c }{THUMOS14} \\
      \cline{4-8}
      \multicolumn{1}{c}{}&\multicolumn{1}{c}{}&\multicolumn{1}{c}{}&0.1&0.2 &0.3 & 0.4 & 0.5  \\
      \hline
      \checkmark &- &- &69.4 &68.1 &64.8 &59.4 &52.3\\
      \checkmark & \checkmark &- &71.5 &70.0 &66.7 &62.0 &55.0\\
      \checkmark & \checkmark & \checkmark &\textbf{73.6}& \textbf{72.1} & \textbf{68.9} & \textbf{62.1} & \textbf{55.9} \\
      \hline
      \end{tabular}}
  \end{center}
  \end{table}

\subsubsection{Effectiveness of the GAF} We also compare the performance of the regular anchor-free detector with and without GAF on two popular datasets. On 
THUMOS14, as Table~\ref{table:4} and Table~\ref{table:5} shows, AFSD, which is equipped with GAF, performs better at most IoU thresholds than the baseline 
model. Table~\ref{table:4} and Table~\ref{table:5} demonstrate that the proposed method is compatible with these methods, which helps 
the AFSD achieve better performance(\textit{e.g.}, 1.8\% improvement at IoU = 0.3 and average 1.0\% mAP improvement over AFSD baseline). Additionally, we compare our method with transformer-based 
method. For fairness, we use the same model I3D as feature extraction(\textit{i.e.}, TadTR~\cite{liu2021end} with I3D), and results(see 
Table~\ref{table:6}) shows that our method performs better than the transformer-based method TadTR~\cite{liu2021end}. 

\begin{table}[t]
  \begin{center}
  \caption{
  % Improvement after adding our model upon AFSD on THUMOS14 and  ActivityNet v1.3.
  Performance of AFSD with our model on THUMOS14.
  }
  \label{table:4}
  \normalsize
  \resizebox{0.45\textwidth}{!}{
    \begin{tabular}{c c c c c c c }
      \hline
      \multicolumn{1}{c}{ \multirow{2}*{Model} }& \multicolumn{6}{c}{THUMOS14}\\
      \cline{2-7}
      \multicolumn{1}{c}{}&0.1&0.2 &0.3 &0.4 &0.5 &Avg.\\
      \hline
      {AFSD (RGB + Flow)} &72.2 &70.8 &67.1 & \textbf{62.2} &55.5 &65.5 \\
      {AFSD (RGB + Flow) + Ours} &\textbf{73.6} &\textbf{72.1} &\textbf{68.9} & 62.1 & \textbf{55.9} & \textbf{66.5}\\
      \hline
      \end{tabular}}
  \end{center}
  \end{table}

\begin{table}[h]
  \begin{center}
  \caption{
  % Improvement after adding our model upon AFSD on THUMOS14 and  ActivityNet v1.3.
  Performance of AFSD with our model on ActivityNet v1.3.
  }
  \label{table:5}
  \normalsize
  \resizebox{0.45\textwidth}{!}{
    \begin{tabular}{c c c c c}
      \hline
      \multicolumn{1}{c}{ \multirow{2}*{Model} } &\multicolumn{4}{c}{ActivityNet v1.3}\\
      \cline{2-5}
      \multicolumn{1}{c}{} &0.5&0.75 &0.95 &Avg.\\
      \hline
      {AFSD (RGB + Flow)}  &52.4 & 35.3 &\textbf{6.5} &31.4\\
      {AFSD (RGB + Flow) + Ours} &\textbf{53.4} &\textbf{36.9} &5.9 &\textbf{32.1}\\
      \hline
      \end{tabular}}
  \end{center}
  \end{table}

  \begin{table}[t]
    \begin{center}
      \caption{
      % Performance comparison with state-of-the-art methods on THUMOS14 and ActivityNet v1.3, measured by mAP at different IoU thresholds on THUMOS14 and ActivityNet v1.3.
      Performance comparison with attention-based method TadTR, using the same backbone I3D.
      }
      \label{table:6}
      \normalsize
      \resizebox{0.45\textwidth}{!}{
      \begin{tabular}{c c c c c c c c }
      \hline
      \multicolumn{1}{ c }{Model} & Backbone & \multicolumn{6}{ c }{THUMOS14} \\
      \cline{3-8}
      \multicolumn{1}{ c }{}& &0.3 & 0.4 & 0.5 & 0.6 & 0.7 & avg.\\
      \hline
      TadTR~\cite{liu2021end} & I3D &62.4& 57.4 & 49.2 & 37.8 & 26.3 & 46.6\\
      \cline{3-8} 
      Ours & I3D &\textbf{68.9}& \textbf{62.1}& \textbf{55.9} & \textbf{44.0} & \textbf{32.4}& \textbf{52.7}\\
      \hline
      \end{tabular}}
      \end{center}
    \end{table}
    
\begin{figure}
  \centering
  \includegraphics[width=0.45\textwidth, height=0.18\textwidth]{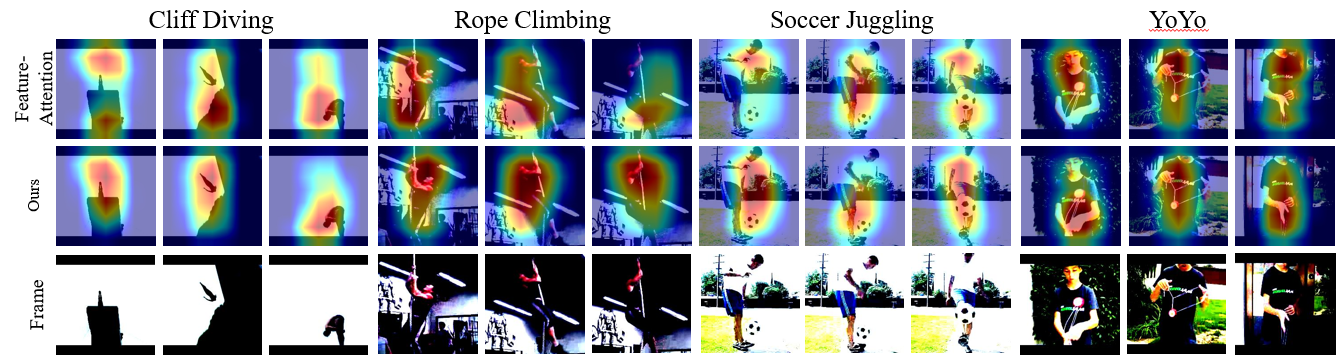}
  \caption{Visualisation results on UCF101 dataset.The final row represents the original feature, while the preceding two rows correspond to different attention-enhanced features.}
  \label{fig:6}
\end{figure}

On ActivityNet v1.3, as shown in Table~\ref{table:4}, we observe that while the performance gains on parts of IoU(\textit{e.g.}, 
0.8\% at IoU = 0.5 and 1.6\% at IoU = 0.75 over AFSD baseline), the performance actually drops in terms of average mAP. The reason 
is that the THUMOS14 dataset has a higher number of action instances per video but the ActivityNet v1.3 does not. The videos in 
the THUMOS14 contain a large amount of non-action instance information and multiple action instances categories, whereas each video 
of ActivityNet v1.3 only contains 1.65 action instances from one or multiple classes on average. An important reason that GAF achieves 
better performance is the intra-relations inside each feature, thus for fewer action instances, as shown in Table~\ref{table:4}, the 
non-attention method actually would be better than our method.

\textbf{Visuilation.} In Fig.~\ref{fig:4}, we visualize four temporal action detection results on videos of THUMOS14. As the four action 
instances show, it can be found that the baseline can't achieve accurate action localization. However, with GAF, the model manages to 
detect action categories more accurately(\textit{e.g.}, example 1) and locate temporal boundaries of the action instances more precisely(\textit{e.g.}, examples 2, 3, 4). 
Additionally, the visual results of action instances and non-action instances are complex and difficult to detect in example 1 and 
example 3. The model with GAF manages to locate the more complete boundaries of action instance than baseline~\cite{lin2021learning}. Meanwhile, as Fig.~\ref{fig:6} shows, we visualized the original feature and different
attention-enhanced features, which proves the features reconstructed by ours outperform the teacher’s features. 

\section{Conclusions}\label{sec:Conclusion}
In this paper, we propose a novel generative attention mechanism designed for the analysis of video action by effectively representing action features within the video stream. Our framework mainly consists of a backbone, GAF and action analyzer. We leverage frame- and segment-level relations within the feature dimension to model action instances across various action-related tasks. Our method demonstrates significant performance gains on various action detection datasets, surpassing other attention-based methodologies by a considerable margin. Additionally, our approach substantiates the effectiveness and progressiveness of modeling action features in video action recognition tasks.
% The results indicate the strength of focusing on feature semantics information as a promising choice for solving TAD tasks.
The results demonstrate that our exploration of feature semantic information effectively improves video action analysis performance.

\bibliographystyle{IEEEtran}
\bibliography{IEEEabrv,egbib}

\begin{thebibliography}{10}
\providecommand{\url}[1]{#1}
\csname url@samestyle\endcsname
\providecommand{\newblock}{\relax}
\providecommand{\bibinfo}[2]{#2}
\providecommand{\BIBentrySTDinterwordspacing}{\spaceskip=0pt\relax}
\providecommand{\BIBentryALTinterwordstretchfactor}{4}
\providecommand{\BIBentryALTinterwordspacing}{\spaceskip=\fontdimen2\font plus
\BIBentryALTinterwordstretchfactor\fontdimen3\font minus \fontdimen4\font\relax}
\providecommand{\BIBforeignlanguage}[2]{{%
\expandafter\ifx\csname l@#1\endcsname\relax
\typeout{** WARNING: IEEEtran.bst: No hyphenation pattern has been}%
\typeout{** loaded for the language `#1'. Using the pattern for}%
\typeout{** the default language instead.}%
\else
\language=\csname l@#1\endcsname
\fi
#2}}
\providecommand{\BIBdecl}{\relax}
\BIBdecl

\bibitem{chen2020internet}
C.~W. Chen, ``Internet of video things: Next-generation iot with visual sensors,'' \emph{IEEE Internet of Things Journal}, vol.~7, pp. 6676--6685, 2020.

\bibitem{ni2024integrated}
H.~Ni, G.~Yu, P.~Chen, B.~Zhou, Y.~Liao, and H.~Li, ``An integrated framework of lateral and longitudinal behavior decision-making for autonomous driving using reinforcement learning,'' \emph{IEEE Transactions on Vehicular Technology}, 2024.

\bibitem{zhang2023cross}
Z.~Zhang, ``Cross-category highlight detection via feature decomposition and modality alignment,'' in \emph{Proceedings of the AAAI Conference on Artificial Intelligence}, vol.~37, 2023, pp. 3525--3533.

\bibitem{wu2022sports}
D.~Wu, H.~Zhao, X.~Bao, and R.~P. Wildes, ``Sports video analysis on large-scale data,'' in \emph{European Conference on Computer Vision}, 2022, pp. 19--36.

\bibitem{Wang2021tdn}
L.~Wang, Z.~Tong, B.~Ji, and G.~Wu, ``Tdn: Temporal difference networks for efficient action recognition,'' in \emph{Proceedings of the IEEE/CVF Conference on Computer Vision and Pattern Recognition}, 2021, pp. 1895--1904.

\bibitem{ouallane2022fusion}
B.~A. Salau, A.~Rawal, and D.~B. Rawat, ``Recent advances in artificial intelligence for wireless internet of things and cyber--physical systems: A comprehensive survey,'' \emph{IEEE Internet of Things Journal}, vol.~9, pp. 12\,916--12\,930, 2022.

\bibitem{liu2022adver}
X.~Zhou, W.~Liang, K.~I.-K. Wang, H.~Wang, L.~T. Yang, and Q.~Jin, ``Deep-learning-enhanced human activity recognition for internet of healthcare things,'' \emph{IEEE Internet of Things Journal}, vol.~7, pp. 6429--6438, 2020.

\bibitem{hussain2020multiview}
T.~Hussain, K.~Muhammad, A.~Ullah, J.~Del~Ser, A.~H. Gandomi, M.~Sajjad, S.~W. Baik, and V.~H.~C. de~Albuquerque, ``Multiview summarization and activity recognition meet edge computing in iot environments,'' \emph{IEEE Internet of Things Journal}, vol.~8, pp. 9634--9644, 2020.

\bibitem{gao2019video}
Z.~Gao, L.~Wang, Q.~Zhang, Z.~Niu, N.~Zheng, and G.~Hua, ``Video imprint segmentation for temporal action detection in untrimmed videos,'' in \emph{Proceedings of the AAAI Conference on Artificial Intelligence}, 2019, pp. 8328--8335.

\bibitem{lin2019bmn}
T.~Lin, X.~Liu, X.~Li, E.~Ding, and S.~Wen, ``Bmn: Boundary-matching network for temporal action proposal generation,'' in \emph{Proceedings of the IEEE/CVF International Conference on Computer Vision}, 2019, pp. 3889--3898.

\bibitem{liu2019multi}
Y.~Liu, L.~Ma, Y.~Zhang, W.~Liu, and S.-F. Chang, ``Multi-granularity generator for temporal action proposal,'' in \emph{Proceedings of the IEEE/CVF Conference on Computer Vision and Pattern Recognition}, 2019, pp. 3604--3613.

\bibitem{ji20123d}
S.~Ji, W.~Xu, M.~Yang, and K.~Yu, ``3d convolutional neural networks for human action recognition,'' \emph{IEEE transactions on pattern analysis and machine intelligence}, vol.~35, pp. 221--231, 2012.

\bibitem{simonyan2014two}
K.~Simonyan and A.~Zisserman, ``Two-stream convolutional networks for action recognition in videos,'' \emph{Advances in neural information processing systems}, vol.~27, 2014.

\bibitem{yang2022recurring}
J.~Yang, X.~Dong, L.~Liu, C.~Zhang, J.~Shen, and D.~Yu, ``Recurring the transformer for video action recognition,'' in \emph{Proceedings of the IEEE/CVF Conference on Computer Vision and Pattern Recognition}, 2022, pp. 14\,063--14\,073.

\bibitem{liu2022graph}
Y.~Liu, H.~Zhang, D.~Xu, and K.~He, ``Graph transformer network with temporal kernel attention for skeleton-based action recognition,'' \emph{Knowledge-Based Systems}, vol. 240, p. 108146, 2022.

\bibitem{chen2019relation}
P.~Chen, C.~Gan, G.~Shen, W.~Huang, R.~Zeng, and M.~Tan, ``Relation attention for temporal action localization,'' \emph{IEEE Transactions on Multimedia}, vol.~22, pp. 2723--2733, 2019.

\bibitem{sridhar2021class}
D.~Sridhar, N.~Quader, S.~Muralidharan, Y.~Li, P.~Dai, and J.~Lu, ``Class semantics-based attention for action detection,'' in \emph{Proceedings of the IEEE/CVF International Conference on Computer Vision}, 2021, pp. 13\,739--13\,748.

\bibitem{wang2023generative}
G.~Wang, P.~Zhao, Y.~Shi, C.~Zhao, and S.~Yang, ``Generative model-based feature knowledge distillation for action recognition,'' in \emph{Proceedings of the AAAI Conference on Artificial Intelligence}, 2024.

\bibitem{xing2023svformer}
Z.~Xing, Q.~Dai, H.~Hu, J.~Chen, Z.~Wu, and Y.-G. Jiang, ``Svformer: Semi-supervised video transformer for action recognition,'' in \emph{Proceedings of the IEEE/CVF Conference on Computer Vision and Pattern Recognition}, 2023, pp. 18\,816--18\,826.

\bibitem{wu2019long}
C.-Y. Wu, C.~Feichtenhofer, H.~Fan, K.~He, P.~Krahenbuhl, and R.~Girshick, ``Long-term feature banks for detailed video understanding,'' in \emph{Proceedings of the IEEE/CVF Conference on Computer Vision and Pattern Recognition}, 2019, pp. 284--293.

\bibitem{zhnag2020ada}
F.~Deng, E.~Jovanov, H.~Song, W.~Shi, Y.~Zhang, and W.~Xu, ``Wildar: Wifi signal-based lightweight deep learning model for human activity recognition,'' \emph{IEEE Internet of Things Journal}, 2023.

\bibitem{quader2020weight}
N.~Quader, M.~M.~I. Bhuiyan, J.~Lu, P.~Dai, and W.~Li, ``Weight excitation: Built-in attention mechanisms in convolutional neural networks,'' in \emph{Proceedings of the European Conference on Computer Vision}, 2020, pp. 87--103.

\bibitem{wang2024paxion}
Z.~Wang, A.~Blume, S.~Li, G.~Liu, J.~Cho, Z.~Tang, M.~Bansal, and H.~Ji, ``Paxion: Patching action knowledge in video-language foundation models,'' \emph{Advances in Neural Information Processing Systems}, vol.~36, 2024.

\bibitem{shi2020weakly}
B.~Shi, Q.~Dai, Y.~Mu, and J.~Wang, ``Weakly-supervised action localization by generative attention modeling,'' in \emph{Proceedings of the IEEE/CVF Conference on Computer Vision and Pattern Recognition}, 2020, pp. 1009--1019.

\bibitem{LIU2022387}
F.~Liu, X.~Xu, X.~Xing, K.~Guo, and L.~Wang, ``Simple-action-guided dictionary learning for complex action recognition,'' \emph{Neurocomputing}, vol. 501, pp. 387--396, 2022.

\bibitem{gao2022pair}
Z.~Gao, L.~Guo, T.~Ren, A.-A. Liu, Z.-Y. Cheng, and S.~Chen, ``Pairwise two-stream convnets for cross-domain action recognition with small data,'' \emph{IEEE Transactions on Neural Networks and Learning Systems}, vol.~33, pp. 1147--1161, 2022.

\bibitem{yuan2016temporal}
J.~Yuan, B.~Ni, X.~Yang, and A.~A. Kassim, ``Temporal action localization with pyramid of score distribution features,'' in \emph{Proceedings of the IEEE/CVF Conference on Computer Vision and Pattern Recognition}, 2016, pp. 3093--3102.

\bibitem{guo2018double}
Y.~Guo, Q.~Wu, C.~Deng, J.~Chen, and M.~Tan, ``Double forward propagation for memorized batch normalization,'' in \emph{Proceedings of the AAAI Conference on Artificial Intelligence}, 2018, pp. 3134--3141.

\bibitem{guo2019nat}
Y.~Guo, Y.~Zheng, M.~Tan, Q.~Chen, J.~Chen, P.~Zhao, and J.~Huang, ``Nat: Neural architecture transformer for accurate and compact architectures,'' \emph{arXiv preprint arXiv:1910.14488}, 2019.

\bibitem{cao2019multi}
J.~Cao, L.~Mo, Y.~Zhang, K.~Jia, C.~Shen, and M.~Tan, ``Multi-marginal wasserstein gan,'' \emph{Advances in Neural Information Processing Systems}, pp. 1776--1786, 2019.

\bibitem{liu2021end}
X.~Liu, Q.~Wang, Y.~Hu, X.~Tang, S.~Bai, and X.~Bai, ``End-to-end temporal action detection with transformer,'' \emph{arXiv preprint arXiv:2106.10271}, 2021.

\bibitem{kong2022human}
Y.~Kong and Y.~Fu, ``Human action recognition and prediction: A survey,'' \emph{International Journal of Computer Vision}, vol. 130, pp. 1366--1401, 2022.

\bibitem{zhou2023learning}
H.~Zhou, Q.~Liu, and Y.~Wang, ``Learning discriminative representations for skeleton based action recognition,'' in \emph{Proceedings of the IEEE/CVF Conference on Computer Vision and Pattern Recognition}, 2023, pp. 10\,608--10\,617.

\bibitem{shi2023tridet}
D.~Shi, Y.~Zhong, Q.~Cao, L.~Ma, J.~Li, and D.~Tao, ``Tridet: Temporal action detection with relative boundary modeling,'' in \emph{Proceedings of the IEEE/CVF Conference on Computer Vision and Pattern Recognition}, 2023, pp. 18\,857--18\,866.

\bibitem{wang2023weakly}
G.~Wang, P.~Zhao, C.~Zhao, S.~Yang, J.~Cheng, L.~Leng, J.~Liao, and Q.~Guo, ``Weakly-supervised action localization by hierarchically-structured latent attention modeling,'' in \emph{Proceedings of the IEEE/CVF International Conference on Computer Vision}, 2023, pp. 10\,203--10\,213.

\bibitem{zhao2023re2tal}
C.~Zhao, S.~Liu, K.~Mangalam, and B.~Ghanem, ``Re2tal: Rewiring pretrained video backbones for reversible temporal action localization,'' in \emph{Proceedings of the IEEE/CVF Conference on Computer Vision and Pattern Recognition}, 2023, pp. 10\,637--10\,647.

\bibitem{goodfellow2014generative}
I.~Goodfellow, J.~Pouget-Abadie, M.~Mirza, B.~Xu, D.~Warde-Farley, S.~Ozair, A.~Courville, and Y.~Bengio, ``Generative adversarial nets,'' in \emph{Proceedings of the Conference on Neural Information Processing Systems}, 2014, pp. 2672--2680.

\bibitem{kingma2013auto}
D.~P. Kingma and M.~Welling, ``Auto-encoding variational bayes,'' \emph{arXiv preprint arXiv:1312.6114}, 2013.

\bibitem{sohn2015learning}
K.~Sohn, H.~Lee, and X.~Yan, ``Learning structured output representation using deep conditional generative models,'' \emph{Advances in neural information processing systems}, pp. 3483--3491, 2015.

\bibitem{bahdanau2014neural}
D.~Bahdanau, K.~Cho, and Y.~Bengio, ``Neural machine translation by jointly learning to align and translate,'' \emph{arXiv preprint arXiv:1409.0473}, 2014.

\bibitem{vaswani2017attention}
A.~Vaswani, N.~Shazeer, N.~Parmar, J.~Uszkoreit, L.~Jones, A.~N. Gomez, {\L}.~Kaiser, and I.~Polosukhin, ``Attention is all you need,'' in \emph{Advances in neural information processing systems}, 2017, pp. 5998--6008.

\bibitem{li2020graph}
J.~Li, X.~Liu, Z.~Zong, W.~Zhao, M.~Zhang, and J.~Song, ``Graph attention based proposal 3d convnets for action detection,'' in \emph{Proceedings of the AAAI Conference on Artificial Intelligence}, 2020, pp. 4626--4633.

\bibitem{lin2021learning}
C.~Lin, C.~Xu, D.~Luo, Y.~Wang, Y.~Tai, C.~Wang, J.~Li, F.~Huang, and Y.~Fu, ``Learning salient boundary feature for anchor-free temporal action localization,'' in \emph{Proceedings of the IEEE/CVF Conference on Computer Vision and Pattern Recognition}, 2021, pp. 3320--3329.

\bibitem{nguyen2019weakly}
P.~X. Nguyen, D.~Ramanan, and C.~C. Fowlkes, ``Weakly-supervised action localization with background modeling,'' in \emph{Proceedings of the IEEE/CVF International Conference on Computer Vision}, 2019, pp. 5502--5511.

\bibitem{soomro2012ucf101}
K.~Soomro, A.~R. Zamir, and M.~Shah, ``Ucf101: A dataset of 101 human actions classes from videos in the wild,'' \emph{arXiv preprint arXiv:1212.0402}, 2012.

\bibitem{THUMOS14}
Y.-G. Jiang, J.~Liu, A.~Roshan~Zamir, G.~Toderici, I.~Laptev, M.~Shah, and R.~Sukthankar, ``{THUMOS} challenge: Action recognition with a large number of classes,'' \url{http://crcv.ucf.edu/THUMOS14/}, 2014.

\bibitem{caba2015activitynet}
B.~G. Fabian Caba~Heilbron, Victor~Escorcia and J.~C. Niebles, ``Activitynet: A large-scale video benchmark for human activity understanding,'' in \emph{Proceedings of the IEEE Conference on Computer Vision and Pattern Recognition}, 2015, pp. 961--970.

\bibitem{lin2018bsn}
T.~Lin, X.~Zhao, H.~Su, C.~Wang, and M.~Yang, ``Bsn: Boundary sensitive network for temporal action proposal generation,'' in \emph{Proceedings of the European Conference on Computer Vision}, 2018, pp. 3--19.

\bibitem{zhao2020bottom}
P.~Zhao, L.~Xie, C.~Ju, Y.~Zhang, Y.~Wang, and Q.~Tian, ``Bottom-up temporal action localization with mutual regularization,'' in \emph{Proceedings of the European Conference on Computer Vision}, 2020, pp. 539--555.

\bibitem{long2019gaussian}
F.~Long, T.~Yao, Z.~Qiu, X.~Tian, J.~Luo, and T.~Mei, ``Gaussian temporal awareness networks for action localization,'' in \emph{Proceedings of the IEEE/CVF Conference on Computer Vision and Pattern Recognition}, 2019, pp. 344--353.

\bibitem{xu2020g}
M.~Xu, C.~Zhao, D.~S. Rojas, A.~Thabet, and B.~Ghanem, ``G-tad: Sub-graph localization for temporal action detection,'' in \emph{Proceedings of the IEEE/CVF Conference on Computer Vision and Pattern Recognition}, 2020, pp. 10\,156--10\,165.

\bibitem{zeng2021graph}
R.~Zeng, W.~Huang, M.~Tan, Y.~Rong, P.~Zhao, J.~Huang, and C.~Gan, ``Graph convolutional module for temporal action localization in videos,'' \emph{IEEE Transactions on Pattern Analysis and Machine Intelligence}, 2021.

\bibitem{zhu2021enriching}
Z.~Zhu, W.~Tang, L.~Wang, N.~Zheng, and G.~Hua, ``Enriching local and global contexts for temporal action localization,'' in \emph{Proceedings of the IEEE/CVF International Conference on Computer Vision}, 2021, pp. 13\,516--13\,525.

\bibitem{brox2009large}
T.~Brox, C.~Bregler, and J.~Malik, ``Large displacement optical flow,'' in \emph{Proceedings of the IEEE/CVF Conference on Computer Vision and Pattern Recognition}, 2009.

\bibitem{kingma2014adam}
D.~P. Kingma and J.~Ba, ``Adam: A method for stochastic optimization,'' \emph{arXiv preprint arXiv:1412.6980}, 2014.

\bibitem{carreira2017quo}
J.~Carreira and A.~Zisserman, ``Quo vadis, action recognition? a new model and the kinetics dataset,'' in \emph{proceedings of the IEEE Conference on Computer Vision and Pattern Recognition}, 2017, pp. 6299--6308.

\end{thebibliography}

\end{document}